\documentclass[letterpaper, 10 pt, conference]{ieeeconf}  

\usepackage{pifont,amssymb,url}
\usepackage{multicol,mathrsfs}
\usepackage[usenames,dvipsnames,svgnames,table]{xcolor}
\usepackage{bm,bbm,mathrsfs,amssymb,pifont,amssymb,pifont,amsfonts,amsmath,stmaryrd,color,amssymb,graphics,graphicx,epstopdf}
\usepackage{amscd,graphicx,array,dsfont,texdraw,tikz,enumerate,multicol,multirow,verbatim}
\usetikzlibrary{arrows,shapes,shadows,positioning,automata,patterns}
\usetikzlibrary{matrix,arrows,calc,trees,decorations.pathmorphing,decorations.markings, decorations.pathreplacing}
\usepackage[utf8x]{inputenc}
\usepackage[T1]{fontenc}
\usepackage{booktabs}

\usetikzlibrary{arrows.meta}
\usetikzlibrary{decorations.pathreplacing}
\newcommand{\tabincell}[2]{\begin{tabular}{@{}#1@{}}#2\end{tabular}}
\renewcommand{\t}{T}

\newcommand{\V}{\mathcal{V}}
\newcommand{\E}{\mathcal{E}}
\newcommand{\G}{\mathcal{G}}
\newcommand{\N}{\mathcal{N}}

\newcommand{\R}{\mathbb{R}}

\newcommand{\col}{\mbox{col}}
\newcommand{\nnum}{\nonumber}

\newcommand{\bremark}{\smallskip\begin{remark}\begin{rm}}
\newcommand{\eremark}{\end{rm}\hfill$\boxempty$\end{remark}\smallskip}
\newcommand{\btheorem}{\smallskip\begin{theorem} \begin{it}}
\newcommand{\etheorem}{\end{it}\end{theorem}\smallskip}
\newcommand{\blemma}{\smallskip\begin{lemma} \begin{it} }
\newcommand{\elemma}{\end{it}\end{lemma}\smallskip}
\newcommand{\bcorollary}{\smallskip\begin{corollary} \begin{it} }
\newcommand{\ecorollary}{\end{it}\end{corollary}\smallskip}
\newcommand{\bdefinition}{\smallskip\begin{definition}\begin{rm}}
\newcommand{\edefinition}{\end{rm}\hfill$\boxempty$\end{definition}\smallskip}
\newcommand{\bproposition}{\smallskip\begin{proposition}\begin{it}}
\newcommand{\eproposition}{\end{it}\end{proposition}\smallskip }
\newcommand{\bexample}{\smallskip\begin{example}\begin{rm}}
\newcommand{\eexample}{\end{rm}\hfill$\boxempty$\end{example}\smallskip}
\newcommand{\bproblem}{\smallskip\begin{problem}\begin{rm}}
\newcommand{\eproblem}{\end{rm}\hfill$\boxempty$\end{problem}\smallskip}

\newcommand{\bassume}{\smallskip\begin{assumption}\begin{rm}}
\newcommand{\eassume}{ \end{rm}\hfill$\boxempty$\end{assumption}\smallskip}
\newcommand{\bfact}{\smallskip\begin{fact}\begin{it}}
\newcommand{\efact}{ \end{it} \end{fact}\smallskip}
\newcommand{\bcondition}{\begin{condition}\begin{rm}}
\newcommand{\econdition}{ \end{rm}\hfill$\boxempty$\end{condition}\smallskip}

\newtheorem{theorem}{Theorem}[section]
\newtheorem{lemma}{Lemma}[section]
\newtheorem{corollary}{Corollary}[section]
\newtheorem{definition}{Definition}[section]
\newtheorem{proposition}{Proposition}[section]
\newtheorem{problem}{Problem}[section]

\newtheorem{myremark}{Remark}[section]
\newenvironment{remark}{\begin{myremark}\normalfont}{\end{myremark}}
\newtheorem{myexample}{Example}[section]
\newenvironment{example}{\begin{myexample}\normalfont}{\end{myexample}}
\newtheorem{assumption}{Assumption}
\newtheorem{fact}{Fact}[section]
\newtheorem{condition}{\it Condition}[section]

\newcommand{\bbm}[1]{\left[\begin{matrix} #1 \end{matrix}\right]}

\newcommand{\bproof}{ \textit{Proof:} \begin{rm} }
\newcommand{\eproof}{ \end{rm} \hfill Q.E.D.}  

\IEEEoverridecommandlockouts                              

\title{\LARGE \bf
Distributed formation control of manipulators' end-effector with internal model-based disturbance rejection
}

\author{Haiwen Wu, Bayu Jayawardhana, Hector Garcia de Marina and Dabo Xu 
\thanks{Haiwen Wu and Bayu Jayawardhana are with Engineering and Technology Institute Groningen, Faculty of Science and Engineering, University of Groningen, Groningen 9747 AG, The Netherlands (e-mails: haiwen.wu@rug.nl; b.jayawardhana@rug.nl).
H.G. de Marina is with Department of Computer Architecture and Automatic Control, Faculty of Physics, Universidad Complutense de Madrid, 28040 Madrid, Spain (email: hgarciad@ucm.es). The work of H.G. de Marina is supported by the grant Atraccion de Talento 2019-T2/TIC-13503 from the Goverment of Madrid.
Dabo Xu is with the School of Automation, Nanjing University of Science and Technology, Nanjing 210094, China (e-mail: dxu@njust.edu.cn).
}
}

\begin{document}

\maketitle
\thispagestyle{empty}
\pagestyle{empty}

\begin{abstract}
This paper addresses the problem of end-effector formation control for manipulators that are subjected to external disturbances: input disturbance torques and disturbance forces at each end-effector. The disturbances are assumed to be non-vanishing and are superposition of finite number of sinusoidal and step signals. The formation control objective is achieved by assigning virtual springs between end-effectors, by adding damping terms at joints, and by incorporating internal model-based dynamic compensators to counteract the effect of the disturbances; all of which presents a clear physical interpretation of the proposed approach. Simulation results are presented to illustrate the effectiveness of the proposed approach.
\end{abstract}

\section{INTRODUCTION}

For the past decade, there have been significant interests in the deployment of multi-robot or autonomous systems that can collectively carry out complex tasks. In this context, distributed formation control plays a key role in achieving and keeping a prescribed formation shape that is necessary to complete higher-level tasks, such as, carrying large payload, search and rescue mission, or environmental monitoring, among others.

Distributed formation control generally aims at controlling a group of robots to achieve a desired geometrical pattern based on the use of local information from on-board sensors. We refer interested readers to a thorough survey in \cite{Oh2015survey}.
In literature, there are a number of different distributed formation control based on the underlying dynamics of the robots/agents. When the agent is considered as a kinematic point (or point mass) whose dynamics is given by single-integrators and double-integrators, simple gradient-based control laws have been proposed and studied, for instance, in
\cite{chan2021,Oh2014distance,Marina2015controlling,Marina2018taming,zhao2015bearing} where different sources of local information (relative position, distance, bearing or vision) are used. The extension of the formation control to other classes of nonlinear systems includes the formation of non-holonomic wheeled robots \cite{Jafarian2016disturbance,vos2016formation}, spacecraft formation flying \cite{Scharf2003survey,Scharf2004survey}, and dynamic positioning of multiple offshore vessels \cite{Xu2017formation}. In all these works, the control input acts directly on the state variables that define the formation. It remains an open problem on the design of formation control for systems where the control input does not act directly on the formation error variables, such as, the formation control of underactuated systems or  end-effector manipulator systems.

In this paper, we investigate the problem of distributed formation control of manipulators' end-effector subjected to external disturbances. In particular we tackle the following two main challenges in this problem.

Firstly, the formation control problem for end-effectors itself is not trivial because the desired formation shape is formed by the end-effectors while the control inputs or the actuators act at the joints' level, which makes this problem challenging. One can consider two level of controllers where distributed formation control law is designed for the formation keeping of end-effectors as kinematic points, and subsequently, the computed velocity at each end-effector for maintaining the formation is back-propagated to the control inputs at the joints' level via inverse kinematics. This multi-level control scheme is, in practice, non-trivial since there is no time-scale separation in the the use of collaborative manipulators for the high-speed robotization in industry, and the computation of inverse kinematics is computationally demanding.

Secondly, based on  existing literature on the disturbances rejection using internal-model-based approach for Euler-Lagrange systems (for example, \cite{Chen1997adaptive,Bayu2008,Lu2019,Wu2019ICCA}), we also consider the compensation of disturbance forces at each end-effector while maintaining the formation. In this case, the disturbances may come from  higher-level tasks such as grasping an objective of unknown load \cite{Verginis2019cooperative,Ren2020fully,Dohmann2020distributed}. We assume that both input disturbance torques and disturbance forces at each end-effector are non-vanishing, and they are assumed to be a  superposition of finite number of sinusoidal and step signals. The incorporation of internal models in our formation control will allow us to both compensate these disturbances and to achieve desired formation shape simultaneously.

In our main result, we present the design of a distance-based distributed formation control of end-effectors based only on local information that is combined with internal-model based compensators to reject the disturbances. 
Our proposed distributed formation control law uses local information that comes from on-board sensor systems defined on local coordinate frame. In other words, the relative information of an end-effector's position with respect to its neighbors and the joints' position/velocity of the robot is independent of its neighbors' frames. The proposed controller is composed of three main components. Firstly, we assign virtual springs between end-effectors. Secondly, we add damping terms at joints. Lastly, we incorporate internal model-based dynamic compensators to counteract the effect of external disturbance. This physics-based control design approach allows us to obtain 
physical interpretation of the proposed approach.
The stability and convergence analysis show that the manipulators' end-effectors converge to the desired formation shape in spite of the presence of external disturbances.

\textit{Notation.}
For a matrix $A\in\R^{m\times n}$, $A^{\t}$ denotes its transpose.
For column vectors $x_{1},\dots,x_{n}$, where $x\in\mathbb{R}^m$, we write $\col(x_{1},\dots,x_{n}) := [x_{1}^{\t},\dots,x_{n}^{\t}]^{\t}$ as the column stacking vector. 
We define the short-hand notation $\overline B := B \otimes I_m$ where $I$ is the identity matrix.

\section{Formulation}\label{sec-form}
\subsection{Manipulator dynamics and kinematics}
Consider a group of $n$-DOF fully-actuated rigid robotic manipulator modeled by 
\cite{Slotine-book,Murray1994,Spong-book}
\begin{equation}\label{sys-dyna}
\begin{aligned}
H_{i}(q_{i})\ddot{q}_{i} + C_{i}(q_{i},\dot{q}_{i})\dot{q}_{i} + g_{i}(q_{i})
= u_{i} + d_{i}
\end{aligned}
\end{equation}
for $i\in\{1,\ldots,N\}$, where $q_{i}(t),\dot{q}_{i}(t),\ddot{q}_{i}(t)\in\R^{n}$ are the generalized joint position, velocity, and acceleration, respectively, $u_{i}(t)\in\R^{n}$ is the generalized joint control forces, $d_{i}(t)\in\R^{n}$ is the external disturbance, $H_{i}(q_{i})\in\R^{n \times n}$ is the inertia matrix, $C_{i}(q_{i},\dot{q}_{i})\in\R^{n \times n}$ is the Coriolis and centrifugal force matrix-valued function, and $g_{i}(q_{i})\in\R^{n}$ is the gravitational torque.

Let $x_{i}(t)\in\R^{m}$ be the $i$th manipulator end-effector position in the task-space (e.g., Cartesian space with $m\in\{2,3\}$) with respect to world frame $\Sigma_{g}$ and $m\leq n$.  The end-effector position can be mapped to its generalized joint displacement via a nonlinear forward kinematics mapping \cite{Murray1994,Spong-book}
\begin{equation}\label{sys-kine}
x_{i} = h_{i}(q_{i}) + x_{i0}
\end{equation}
where $h_{i}:\R^{n}\to\R^{m}$ is the mapping from joint-space to task-space, and $x_{i0}\in\R^{m}$ is the position of manipulator base with respect to the world frame $\Sigma_{g}$.

Differentiating \eqref{sys-kine} with respect to time gives the relation between the task-space velocity and joint velocity \cite[pp.~196]{Murray1994}, \cite[pp.~122]{Spong-book}
\begin{equation}\label{sys-J}
\dot{x}_{i} = J_{i}(q_{i})\dot{q}_{i},~~ J_{i}(q_{i}) := \frac{\partial h_{i}(q_{i})}{\partial q_{i}}
\end{equation}
where $J_{i}(q_{i})\in\R^{m\times n}$ is the Jacobian matrix of the forward kinematics.

As will be defined precisely later in the control problem formulation, we are interested in the distributed formation control of end-manipulators which are also interacted with the dynamic environment. For representing this dynamic interaction, we consider the presence of external disturbance $d_i$ in each manipulator that can be decomposed as
\begin{equation}\label{defn-d}
d_{i} = d_{M,i} + J_{i}^{T}(q_{i})d_{E,i}
\end{equation}
where $d_{M,i}(t)\in\R^{n}$ is input disturbance and $d_{E,i}(t)\in\R^{m}$ is the external force at end-effector. For simplicity of control design and analysis, we assume that both of them are generated by the following exosystems
\begin{equation}
\label{exosystem}
\begin{aligned}\dot{v}_{M,i} &= S_{M,i}v_{M,i}  \\
 d_{M,i} &= G_{M,i}v_{i}
\end{aligned}
\text{ and }\,\,
\begin{aligned}\dot{v}_{E,i} &= S_{E,i}v_{E,i}  \\
 d_{E,i} &= G_{E,i}v_{E,i}
\end{aligned}
\end{equation}
for $i=1,\dots,N$, with states $v_{M,i}$ and $v_{E,i}$ of appropriate dimensions.



We assume that the exosystems $S_{M,i}$ and $S_{E,i}$ in \eqref{exosystem} are neutrally stable, e.g., all the eigenvalues of matrices $S_{M,i}$ and $S_{E,i}$ are distinct (in general) and lie on the imaginary axis, respectively.

Throughout this paper, we assume standard properties on the inertia and Coriolis matrices $H_{i}$ and $C_{i}$ that are commonly inherited in most Euler-Lagrange (EL) systems  \cite{Ortega-book,Kelly-book}. In particular, we assume the following properties.  
\begin{enumerate}[{\bf P1}]
  \item The inertia matrix $H_{i}(q_{i})$ is positive definite. More specifically, there exist $c_{i,\rm min},c_{i,\rm max}>0$ such that
      \begin{align*}
      c_{i,\rm min}I \leq H_{i}(q_{i}) \leq c_{i,\rm max}I,~~ \forall q_{i}\in\R^{n}.
      \end{align*}

  \item The matrix-valued function $\dot{H}_{i}(q_{i},\dot{q}_{i})-2C_{i}(q_{i},\dot{q}_{i})$ is skew symmetric, i.e., for any differentiable function $q_{i}(t)\in\R^{n}$ and its time derivative $\dot{q}_{i}(t)$,
      \begin{equation}\label{Property2}
      \dot{H}_{i}(q_{i},\dot{q}_{i}) = C_{i}(q_{i},\dot{q}_{i}) + C_{i}^{\t}(q_{i},\dot{q}_{i})
      \end{equation}
      where $\dot{H}_{i}(q_{i},\dot{q}_{i}) = \sum_{j=1}^{n}\frac{\partial H_{i}}{\partial q_{ij}}\dot{q}_{ij}$.



\end{enumerate}

\medskip


\subsection{Graph on formation}
Let $1 < N  \in\mathbb{N}$ define the number of robotic manipulators whose end-effectors must maintain a specific formation. The neighboring relationships between their end-effectors are described by an undirected and connected graph $\G := \{\V,\E\}$ with the vertex set $\V := \{1,\cdots,N\}$ and the ordered edge set $\E\subset\V\times\V$. The set of the neighbors for the end-effector $i$ is given by $\N_{i} := \{j\in\V:(i,j)\in\E\}$. We use $|\V|=N$ and $|\E|$ to denote the number of vertices and edges of $\G$, respectively. We define the elements of the incidence matrix $B\in\R^{|\V|\times|\E|}$ of $\G$ by
\begin{align*}
b_{ik} = \left\{
\begin{array}{ll} +1, & i=\E_{k}^{\text{tail}} 
\\
-1, &  i=\E_{k}^{\text{head}} 
\\
0, & \text{otherwise} \end{array}
\right.
\end{align*}
where $\E_{k}^{\text{tail}}$ and $\E_{k}^{\text{head}}$ denote the tail and head nodes, respectively, of the edge $\E_{k}$, i.e., $\E_{k}=(\E_{k}^{\text{tail}}, \E_{k}^{\text{head}})$.
Note that $B^{T}\mathbf{1}_{|\V|} = 0$, where $\mathbf{1}_n\in\mathbb{R}^n$ is the vector whose all elements are ones.

\subsection{End-effector distributed formation control problem}
We refer to \emph{configuration} as the stacked vector of end-effectors' positions $x=\col(x_{1},\ldots,x_{N})\in\R^{mN}$, and we refer to \emph{framework} as the pair $(\G,x)$. Given a \emph{reference configuration} $x^*$, we define the \emph{desired shape} as the set
\begin{equation}
    \mathcal{S}:=\{x : x=(I_N \otimes R) x^* + \mathbf{1}_N \otimes b, \, R\in\text{SO($m$)}, b\in\mathbb{R}^m \}.
    \label{eq: ds}
\end{equation}
Let us stack all joint coordinates into $q=\col(q_{1},\ldots,q_{N})$ and $\dot{q} = \col(\dot{q}_{1},\ldots,\dot{q}_{N})$. Note that $\mathcal{S}$ accounts for any arbitrary translation and rotation. However, the \emph{working space} for the end-effectors is constrained since the bases of the arm manipulators are fixed. Therefore, we define $\mathcal{S}_W \subset \mathcal{S}$ as the subset of shapes that are both desired and reachable by the end-effectors.

We are now ready to formulate our formation control problem of end-effectors as follows.
\begin{problem}\label{prob}
\emph{(End-effector distributed formation control problem)}
For a group of $N$ manipulators given by \eqref{sys-dyna}, whose end-effector positions are as in \eqref{sys-kine}, 
design a distributed control law of the form
\begin{equation}
\left\{
\begin{aligned}
\dot{\chi}_{i} &= f_{ci}\Big((x_i - x_j),q_{i},\dot{q}_{i},\chi_{i}\Big) \\
u_{i} &= h_{ci}\Big((x_i - x_j),q_{i},\dot{q}_{i},\chi_{i}\Big)
\end{aligned}
\right. , \quad j\in\mathcal{N}_i.
\label{prob-law}
\end{equation}
such that $x(t) \to \mathcal{S}_W$ and $\dot{q}(t) \to \mathbf{0}$ as $t\to\infty$ for initial conditions $x(0)$ that start in a neighborhood of $\mathcal{S}_W$. The state $\chi_{i}(t)$ in (\ref{prob-law}) is the compensator state which will be designed later.
\end{problem}\vspace{0.1cm}

In this paper, we will focus on the distributed control design framework where we can directly extend the well-known distributed formation control of mobile robots (modeled as single-integrator agents) to the formation control of end-effectors in arm manipulators. In the latter case, the dynamics is given by second-order systems as in \eqref{sys-dyna} while the control input is defined at the joint level.

In order to illustrate our design framework, we consider the use of displacement-based \cite{de2020maneuvering} and distance-based distributed formation control \cite{de2016distributed}, which are two of well-studied distributed control methods. We note that our proposed framework is extensible to other gradient-descent based approaches, such as the bearing-rigidity \cite{zhao2015bearing}.

For the displacement-based formation control, we have that $R = I_m$ in (\ref{eq: ds}). In other words, it only admits desired formation shapes which are given by the translation of $x^*$. On the other hand, the distance-based formation control admits desired formation shapes that are both the translation and rotation of $x^*$. 

The shape displayed by the reference configuration $x^*$ can also be described by a set of geometric relations between the neighboring end-effectors. If $\G$ is connected, then the relative positions defined by the graph $z^* = \overline B^T x^*$ define uniquely the desired shape in displacement-based control, i.e., we have the singleton $\mathcal{Z}_{\text{displacement}} := \{z : z = z^*\}$. Note that the elements of $z^*$ correspond to the ordered $z^*_{ij} = z^*_k := x^*_i - x_j^*, \, (i,j) = \mathcal{E}_k\in\mathcal{E}$. If $\G$ is infinitesimally minimally rigid (e.g., it has a minimum number of edges for being infinitesimally rigid \cite{AnYuFiHe08}), then the set of distances $\|z_{ij}^*\|, (i,j)\in\mathcal{E}$ define locally\footnote{In the sense that it might define a finite number of other shapes.} the desired shape, i.e., we have the set $\mathcal{Z}_{\text{distance}} := \{z : \|z_{ij}\| = \|x_i - x_j\|, \, (i,j)\in\mathcal{E}\}$.

There are some advantages and disadvantages between the use of displacement-based and distance-based formation control. The former requires a minimum number of edges for $\mathcal{G}$, and the resultant control action for pure kinematic agents is linear. However, the desired shape can only be a translation version of $x^*$, and the algorithm require neighboring agents to share the same frame of coordinates to control the common vector $z_{ij}^*$. On the other hand, the distance-based formation control requires more edges, e.g., at least $(2N - 3)$ in 2D, and the control action for pure kinematic agents is nonlinear leading to only local stability around $\mathcal{S}$. Nevertheless, it allows a more flexible $\mathcal{S}$, e.g., it allows rotations for $x^*$ and the agents do not need to share a common frame of coordinates since they are controlling the scalars $\|z_{ij}^*\|$.

\begin{remark}
Industrial manipulators commonly use a spherical wrist at the end-effector, and therefore they can achieve any desired orientation at a given end-effector's position \cite[pp.~95]{Murray1994}. This allows us to focus only on the position of the end-effector since their orientation is \emph{decoupled} thanks to the spherical wrist.
\end{remark}

\begin{figure*}[ht]
  \centering
  \includegraphics[width=0.95\textwidth]{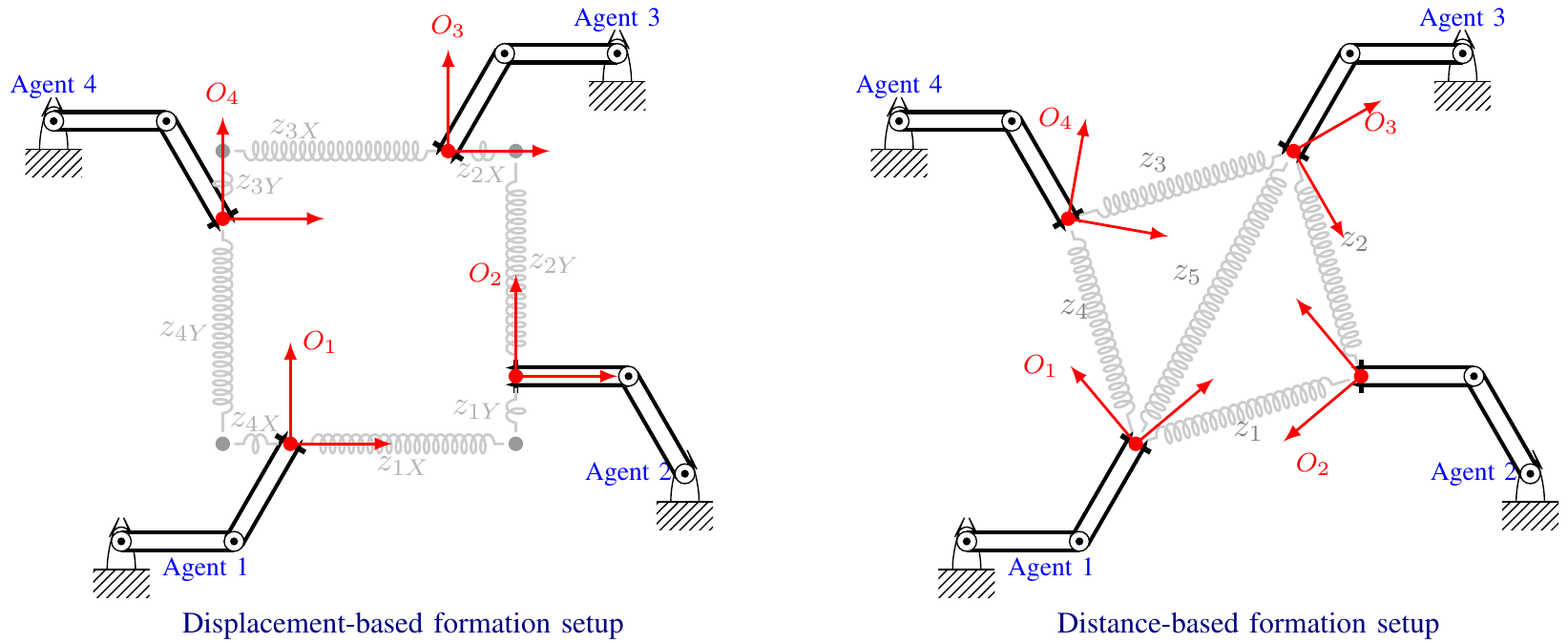}
\caption{End-effector formation of 4 two-link planar manipulators in  the horizontal plane. Left: displacement-based formation control. Right: distance-based formation control. The springs in gray color are virtual couplings assigned on the graph of the end-effectors. $O_{1}$ to $O_{4}$ in red color are local coordinate systems attached at each end-effector. For the displacement-based setup, the orientations of local coordinate systems need to be aligned with global coordinate system. For the distance-based setup, the orientations of local coordinate systems do not necessarily share a common coordinate system.} \label{fig-robots}
\end{figure*}

\section{Control Design}\label{sec-main}
We propose a distributed control design framework where each 
controller $u_{i}$ comprises of three main elements: an end-effector formation controller $u_{i}^{f}$, a joint velocity controller $u_{i}^{v}$, and $u_{i}^{d}$ to reject disturbance. In Section~\ref{subSec-formation}, we design control law $u_{i}^{f}$ by using virtual spring. Subsequently, in Section~\ref{subSec-joint}, we design $u_{i}^{v}$ based on the passivity property between joint torque and joint velocity. Finally, in Section~\ref{subSec-dist}, we design control law $u_{i}^{d}$ based on internal model principle to counteract external disturbances. 

Generally speaking, for solving Problem~\ref{prob}, we firstly employ the virtual spring approach to the end-effectors and introduce standard distributed formation controllers that are based on gradient-descent approach. 
The resulting distributed formation control law defined in the end-effector space is propagated to the joint space via passivity-based approach. The internal-model based compensation of external disturbances can then be designed accordingly.

\subsection{Formation control using virtual spring}\label{subSec-formation}

To achieve the desired formation shape, we start by assigning virtual springs \cite[Chapter~12.2]{van2014pH} on the undirected graph $\G$ of the end-effectors, as depicted in Fig.~\ref{fig-robots}. That is, each edge of $\E$ between the manipulators end-effectors are considered interconnected by virtual couplings, which shape the energy function of the network.
If all of the couplings of the network reach their minimum potential energy, the desired formation is reached.

Consider edge $k$ between agents $i$ and $j$ with virtual coupling. Let us define the following error signal for each edge $k$ of $\mathcal{G}$
\begin{equation}\label{ek}
e_k(t) := f_e(z_k(t), z_k^*),
\end{equation}
where $f_e : \mathbb{R}^m \to \mathbb{R}^l$, and $l\in\mathbb{N}$ will depend on the chosen formation control strategy, e.g., $f_e = \|z_k\|^2 - \|z_k^*\|^2$  for distance-based formation control, and $f_e = z_k - z^*_k$ for displacement-based formation control. Each end-effector in the edge $\mathcal{E}_k = (i,j)$ will apply the gradient descent of $V_k(e_k) = \frac{1}{2} \|e_k\|^2$ as its control input (e.g., its velocity when it is described by kinematic point) in order to reach the minimum of $V$ that coincides with the desired shape. It can be checked that the following equality $\nabla_{x_i} V_k = -\nabla_{x_j} V_k \in\mathbb{R}^m$ is satisfied since $z_{ij} = x_i - x_j$. Let us stack all the $e_k$ in $e\in\mathbb{R}^{l|\mathcal{E}|}$ and define $V(e) := \sum_{k=1}^{|\mathcal{E}|}V_k(e_k)$. For compact representation, let us define the $m$-dimensional agent-wise displacement measurement $\widehat{e}_{i}(t)\in\R^{m}$, $i=1,\ldots,N$ by
\begin{equation}\label{ei}
\widehat{e}_{i} := \nabla_{x_{i}} V(e)
\end{equation}
or equivalently,
\begin{equation}
\widehat{e}_{i} := \sum_{k=1}^{|\E|} b_{ik} D_{k}(z_{k})e_{k}
\end{equation}
where $D_{k}(z_{k}) = \frac{\partial f_{e}(z_{k},z_{k}^{*})}{\partial z_{k}}$, e.g., for the displacement-based formation control: $D_{k}(z_{k}) = 1$ and for the distance-based formation control: $D_{k}(z_{k}) = 2z_{k}$.

Since the virtual springs are assigned between end-effectors, while the actuators are embedded in joints, the corresponding formation control law $u_{i}^{f}$ of agent $i$ can be written as
\begin{equation}\label{u-i}
u_{i}^{f} = - K_{P}J_{i}^{T}(q_{i})\nabla_{x_{i}} V(e) = - K_{P}J_{i}^{T}(q_{i})\widehat{e}_{i}
\end{equation}
with design parameter $K_{P}$, where $J_{i}(q_{i})$ is the manipulator Jacobian matrix and $\widehat{e}_{i}$ is defined in \eqref{ei}.

Let us stack all the $\widehat{e}_{i}$ in $\widehat{e}\in\R^{mN}$, so that we can write it in the following compact form 
\begin{equation}\label{defn-e-wide}
\widehat{e} = \nabla_x V.
\end{equation}
More precisely, for the displacement-based and for the distance-based ones, we have
\begin{align}
\widehat{e}_\text{displacement} &= \nabla_x V_{\text{displacement-based}} = \overline Be_\text{displacement} \\
\widehat{e}_\text{distance} &= \nabla_x V_{\text{distance-based}} = \overline BD(z)e_\text{distance} \label{eq: dist}
\end{align}
where $D(z)=\text{block diag}(D_{1}(z_{1}),\ldots,D_{|\E|}(z_{|\E|}))$.

We note two relevant facts that will be useful for our main technical results. First, $B^TB$ is positive definite if $\mathcal{G}$ does not contain any cycles. Second, $D^{T}(z)\overline{B^TB}D(z)$ is positive definite if $\mathcal{G}$ is infinitesimally and minimally rigid. Roughly speaking, infinitesimally rigid means that all the positions $x_i$ are in a generic configuration, e.g., they are not collinear if $m=2$ or coplanar if $m=3$. Note that if the formation is infinitesimally rigid at $\mathcal{S}$, then it is a neighborhood of $\mathcal{S}$ as well.

\subsection{Joint velocity control}\label{subSec-joint}
For solving the static formation control problem, we proceed by designing a control law to stabilize the joint velocity at origin.
Let us define
\begin{equation}\label{xi}
\xi_{i}(t) := \dot{q}_{i}(t)
\end{equation}
which satisfies, by using \eqref{sys-dyna},
\begin{equation}\label{eq: xi}
\dot{\xi}_{i} = H_{i}^{-1}(q_{i})\Big(u_{i} + d_{i} - C_{i}(q_{i},\xi_{i})\xi_{i} - g_{i}(q_{i})\Big).
\end{equation}

According to the well-known passivity of manipulators from joint torque to joint velocity \cite{Spong-book}, we introduce the following controller, which consists of a damping term and a gravity compensation term
\begin{equation}\label{u-v}
u_{i}^{v} = -K_{D}\xi_{i} + g_{i}(q_{i})
\end{equation}
with a design parameter $K_{D}>0$. Let $V_{i}^{v} = \xi_{i}^{T}H_{i}(q_{i})\xi_{i}$ be the storage function. When $d_{i}$ is neglected, it can be checked that the time derivative $\dot{V}_{i}^{v}$ satisfies
\begin{align*}
\dot{V}_{i}^{v}|_{\eqref{eq: xi}+\eqref{u-v}} = -\xi_{i}^{T}K_{D}\xi_{i}.
\end{align*}

\subsection{Disturbance rejection}\label{subSec-dist}
This part aims at dealing with the disturbance modeled by \eqref{exosystem}. 
Inspired by the output regulation theory, internal model-based controllers are implemented to compensate and to reject these disturbances.
\begin{itemize}
  \item To counteract the effect of the $d_{M,i}$ of \eqref{defn-d}, we introduce the following internal model dynamics \cite{Bayu2008,Jafarian2016disturbance}
  \begin{equation}\label{im-I}
  \left\{
  \begin{aligned}
  \dot{\eta}_{i} &= A_{M,i}\eta_{i} - \Gamma_{M,i}^{T}\xi_{i} \\
  \tau_{M,i} &= \Gamma_{M,i}\eta_{i}
  \end{aligned}
  \right. ,~~ i=1,\ldots,N
  \end{equation}
  with state $\eta_{i}(t)\in\R^{l_{i}}$ for some positive integer $l_{i}$, where $A_{M,i} + A_{M,i}^{T} = 0$, $(A_{M,i},\Gamma_{M,i})$ is observable, and there is $\Sigma_{M,i}$ such that
  \begin{align*}
    \Sigma_{M,i} S_{M,i} = A_{M,i} \Sigma_{M,i},~~ \Gamma_{M,i}^{T}\Sigma_{M,i} + G_{M,i} = 0.
  \end{align*}

  \item To counteract the effect of the external force at each end-effector and propagated to its joints: $J_{i}^{T}(q_{i})d_{E,i}$ as in \eqref{defn-d}, we introduce the following internal model dynamics
  \begin{equation}\label{im-II}
  \left\{
  \begin{aligned}
  \dot{\zeta}_{i} &= A_{E,i}\zeta_{i} - \Gamma_{E,i}^{T}J_{i}(q_{i})\xi_{i} \\
  \tau_{E,i} &= J_{i}^{T}(q_{i})\Gamma_{E,i}\zeta_{i}
  \end{aligned}
  \right. ,~~ i=1,\ldots,N
  \end{equation}
  with state $\zeta_{i}(t)\in\R^{\ell_{i}}$ for some positive integer $\ell_{i}$,   where $A_{E,i} + A_{E,i}^{T} = 0$, $(A_{E,i},\Gamma_{E,i})$ is observable, and there is $\Sigma_{E,i}$ such that
  \begin{align*}
    \Sigma_{E,i} S_{E,i} = A_{E,i} \Sigma_{E,i},~~ \Gamma_{E,i}^{T}\Sigma_{E,i} + G_{E,i} = 0.
  \end{align*}

\end{itemize}

Hence, the total disturbance compensator is the sum of $\tau_{M,i}$ and $\tau_{E,i}$ given by
\begin{align*}
u_{i}^{d} =  \Gamma_{M,i}\eta_{i} + J_{i}^{T}(q_{i})\Gamma_{E,i}\zeta_{i}.
\end{align*}
For compactness of presentation, we denote
\begin{align*}
\chi_{i} &= \bbm{\eta_{i} \\ \zeta_{i}},~~
A_{i} = \bbm{A_{M,i} & 0 \\ 0 & A_{E,i}} \\
\Gamma_{i}(q_{i}) &= \bbm{\Gamma_{M,i} & J_{i}^{T}(q_{i})\Gamma_{E,i}}.
\end{align*}
Then the internal models \eqref{im-I} and \eqref{im-II} can be rewritten in the following compact form
\begin{equation}\label{im-all}
\left\{
\begin{aligned}
\dot{\chi}_{i} &= A_{i}\chi_{i} - \Gamma_{i}^{T}(q_{i})\xi_{i}\\
u_{i}^{d} &=  \Gamma_{i}(q_{i})\chi_{i}.
\end{aligned}
\right.
\end{equation}
By defining the following coordinate transformations
\begin{align}\label{IM-error}
\tilde{\chi}_{i} = \bbm{\tilde{\eta}_{i} \\ \tilde{\zeta}_{i}} = \bbm{\eta_{i} - \Sigma_{M,i}v_{M,i} \\ \zeta_{i} - \Sigma_{E,i}v_{E,i}},~~
\tilde{u}_{i}^{d} = u_{i}^{d} - d_{i}
\end{align}
for each $i=1,\ldots,N$, it is straightforward to show that
\begin{equation}\label{cls-dis}
\left\{
\begin{aligned}
\dot{\tilde{\chi}}_{i} &= A_{i}\tilde{\chi}_{i} - \Gamma_{i}^{T}(q_{i})\xi_{i}\\
\tilde{u}_{i}^{d} &=  \Gamma_{i}(q_{i})\tilde{\chi}_{i}.
\end{aligned}
\right.
\end{equation}
Using storage function $V_{i}^{d} = \frac{1}{2}\tilde{\chi}_{i}^{T}\tilde{\chi}_{i}$, it follows immediately that
\begin{align*}
\dot{V}_{i}^{d}|_{\eqref{cls-dis}} &= \tilde{\chi}_{i}^{T}[A_{i}\tilde{\chi}_{i} - \Gamma_{i}^{T}(q_{i})\xi_{i}] = -\xi_{i}^{T}\tilde{u}_{i}^{d}.
\end{align*}
This implies that system \eqref{cls-dis} is lossless with respect to input $\xi_{i}$ and output $\tilde{u}_{i}^{d}$.

\section{Main Result}\label{subsec-main}
In this section, we will combine the individual control laws 
$u_{i}^{f}$, $u_{i}^{v}$ and $u_{i}^{d}$ above and analyze the solvability of Problem \ref{prob} in the following theorem.

Before presenting the main result, we need the following assumption on the Jacobian matrix $J(q) = \text{block diag}(J_{1}(q_{i}),\dots,J_{N}(q_{N}))$, which is standard in manipulator task-space control \cite{Cheah2003approximate,Dixon2007adaptive,Wang2020dynamic}.
\begin{assumption}\label{assume-J}
The Jacobian matrix $J(q)$ is full rank in a neighborhood $\mathcal{S}_{W_r}$ of $\mathcal{S}_W$, i.e., for some positive constant $r\in\mathbb{R}_{>0}$ we define the set $\mathcal{S}_{W_{r}} := \{x\in\R^{mN}: \|e\| < r \}$.
\end{assumption}\vspace{0.1cm}

Note that this assumption is not very restrictive. If the desired shape has been designed such that $J(q)$ is full rank in the workspace, then by continuity argument, it is clear that the Jacobian will still be full rank in a neighborhood of $\mathcal{S}_W$.

\begin{theorem}\label{thm-I}
Consider $N$ robot manipulators \eqref{sys-dyna} with undirected  graph $\G$ for the formation of end-effectors and with known systems parameters. 
Then the end-effector formation control problem can be solved locally starting from a configuration $x(0)\in \mathcal{S}_{W_r}$, by the following distributed formation control law (in the form of \eqref{prob-law})
\begin{equation}\label{law-I}
\left\{
\begin{aligned}
u_{i} &= - K_{P}J_{i}^{T}(q_{i}) \widehat{e}_{i} - K_{D}\xi_{i} + g_{i}(q_{i}) + \Gamma_{i}^{T}(q_{i})\chi_{i} \\
\dot{\chi}_{i} &= A_{i} - \Gamma_{i}(q_{i})\xi
\end{aligned}
\right.
\end{equation}
with matrices $A_{i}$, $\Gamma_{i}(q_{i})$ given by
\begin{equation*}
A_{i} = \bbm{A_{M,i} & 0 \\ 0 & A_{E,i}},~~
\Gamma_{i}(q_{i}) = \bbm{\Gamma_{M,i} & J_{i}^{T}(q_{i})\Gamma_{E,i}}
\end{equation*}
for all $i=1,\ldots,N$, where constant parameters $K_{P}>0$, $K_{D}>0$, the pairs $(A_{M,i},\Gamma_{M,i})$, $(A_{E,i},\Gamma_{E,i})$ are specified in \eqref{im-I}, \eqref{im-II} with the corresponding assumptions, respectively, and vectors $\widehat{e}_{i}$, $\xi_{i}$ are given in \eqref{ei}, \eqref{xi}, respectively.
\end{theorem}\vspace{0.1cm}

\bproof
Substituting control law \eqref{law-I} into  \eqref{sys-dyna} and using coordinate transformations \eqref{ek}, \eqref{IM-error} gives the following closed-loop error system in the compact form
\begin{equation}\label{cls-I}
\left\{
\begin{aligned}
\dot{q} &= \xi \\
\dot{\tilde{\chi}} &= A\tilde{\chi} - \Gamma^{T}(q)\xi \\
\dot{e} &= D^{T}(z)\overline{B}^{T}J(q)\xi \\
\dot{\xi} &= H^{-1}(q)\Big(-K_{P}J^{T}(q)\widehat{e} - K_{D}\xi - C(q,\xi)\xi \\
&\quad + \Gamma(q)\tilde{\chi} \Big)
\end{aligned}
\right.
\end{equation}
where $q$, $\tilde{\chi}$, $\widehat{e}$, $\xi$, $g(q)$ are the stacked vectors of  $q_{i}$, $\tilde{\chi}_{i}$, $\widehat{e}_{i}$, $\xi_{i}$, $g_{i}(q_{i})$, respectively, for all $i=1,\ldots,N$, and $H(q)$, $C(q,\xi)$, $J(q)$, $A$, $\Gamma(q)$ are the block diagonal matrices of $H_{i}(q_{i})$, $C_{i}(q_{i},\xi_{i})$, $J_{i}(q_{i})$, $A_{i}$, $\Gamma_{i}(q_{i})$, respectively, for all $i=1,\ldots,N$.
Define a Lyapunov function candidate $U := U(q,\tilde{\chi},e,\xi)$ for \eqref{cls-I} by
\begin{align*}
U = \frac{1}{2}\tilde{\chi}^{T}\tilde{\chi}  + \frac{1}{2}e^{T}K_{P}e + \frac{1}{2}\xi^{T}H(q)\xi.
\end{align*}
Its time derivative satisfies
\begin{align*}
\dot{U}|_{\eqref{cls-I}} &= \tilde{\chi}^{T}\big(A\tilde{\chi} - \Gamma^{T}(q)\xi\big) + e^{T}K_{P}D^{T}(z)\overline{B}^{T}J(q)\xi \\
&\quad + \xi^{T}\Big(-K_{P}J^{T}(q)\widehat{e}  - K_{D}\xi + \Gamma(q) \tilde{\chi} \Big)
\end{align*}
Since $A$ is a skew-symmetric matrix, and $\widehat{e} = \overline{B}D(z)e$ as defined in \eqref{ei}, we have
\begin{align*}
\dot{U}|_{\eqref{cls-I}} = -\xi^{T}K_{D}\xi.
\end{align*}
In order to show the asymptotic stability of the desired shape, we can now invoke La-Salle's invariance principle 
\cite[pp.~128]{Khalil2002}. Toward this end, we need to show the largest invariant in
\begin{equation*}
\{(q,\tilde{\chi},e,\xi): \dot{U} = 0 \} = \{(q,\tilde{\chi},e,\xi): \xi = \mathbf{0} \}.
\end{equation*}
Substituting $\xi = \mathbf{0}$ into \eqref{cls-I} gives the following dynamics
\begin{equation}\label{cls-II}
\left\{
\begin{aligned}
\dot{q} &= \mathbf{0} \\
\dot{\tilde{\chi}} &= A\tilde{\chi}  \\
\dot{e} &= \mathbf{0} \\
\mathbf{0} &= -K_{P}J^{T}(q)\widehat{e}  + \Gamma(q)\tilde{\chi}
\end{aligned}
\right. .
\end{equation}
The time derivative of the last equality in \eqref{cls-II} satisfies
\begin{align*}
\mathbf{0} &= -K_{P}\left[\dot{J}^{T}(q)\widehat{e} + J^{T}(q)\dot{\widehat{e}}\right]  + \dot{\Gamma}(q) \tilde{\chi} + \Gamma(q)\dot{\tilde{\chi}}
\end{align*}
where 
\begin{align*}
\dot{J}(q) = \sum_{i=1}^{N}\sum_{j=1}^{n}\frac{\partial J}{\partial q_{ij}}\dot{q}_{ij},~~ \dot{\Gamma}(q) = \sum_{i=1}^{N}\sum_{j=1}^{n} \frac{\partial \Gamma }{\partial q_{ij}}\dot{q}_{ij}.
\end{align*}
Since $\dot{q} = \mathbf{0}$ and $\dot{\widehat{e}} = \mathbf{0}$ in the invariant set, we obtain
\begin{equation*}
\mathbf{0} = \Gamma(q)\dot{\tilde{\chi}}.
\end{equation*}
For the above, using the second equality of \eqref{cls-II} gives
\begin{equation}\label{obs-II}
\mathbf{0} = \Gamma(q) A \tilde{\chi}.
\end{equation}
Following the similar computation, repeated time derivative of \eqref{obs-II} $p-1$ times where $p$ is the dimension of matrix $A$, we have that
\begin{equation}\label{zeros}
\left\{
\begin{aligned}
\mathbf{0} &= \Gamma(q) A^{2} \tilde{\chi} \\
&\vdots \\
\mathbf{0} &= \Gamma(q) A^{p} \tilde{\chi}
\end{aligned}
\right.
\end{equation}
holds. By invoking the Cayley–Hamilton theorem, there is a set of real numbers $\{\alpha_{i}\}_{i=1}^{N}$ such that
\begin{equation}\label{C-H}
A^{p} + \alpha_{1}A^{p-1} + \dots + \alpha_{p-1}A + \alpha_{p}I = 0.
\end{equation}
By using \eqref{C-H}, we have
\begin{align}\label{zero}
&\Gamma(q)\tilde{\chi} \nnum\\
&= -\frac{1}{\alpha_{p}}\Gamma(q) \left(A^{p} + \alpha_{1}A^{p-1}  + \dots + \alpha_{p-1}A\right) \tilde{\chi} \nnum\\
&= -\frac{1}{\alpha_{p}}\Gamma(q) A^{p}\tilde{\chi}  -\frac{\alpha_{1}}{\alpha_{p}}\Gamma(q) A^{p-1}\tilde{\chi} - \dots -\frac{\alpha_{p-1}}{\alpha_{p}}\Gamma(q) A\tilde{\chi} \nnum\\
& = \mathbf{0}
\end{align}
where the last equality is due to \eqref{obs-II} and \eqref{zeros}.
Substituting \eqref{zero} into \eqref{cls-II} results in $J^{T}(q)\widehat{e} = \mathbf{0}$.
This implies that $\widehat{e} = \mathbf{0}$ as long as $J(q)$ is nonsingular in a neighborhood of $\mathcal{S}_W$. Also, note that in the case of distance-based control, the matrix $D^T(z)\overline B^T$ is full rank in a neighborhood of $\mathcal{S}_W$.
Therefore, $\widehat{e} = \mathbf{0}$ immediately implies $e=\mathbf{0}$ in a neighborhood of $\mathcal{S}_W$. Hence, within $\mathcal{S}_W$, the largest invariant set with respect to \eqref{cls-I} in
\begin{equation*}
\{(q,\tilde{\chi},e,\xi): \dot{U} = 0 \} = \{(q,\tilde{\chi},e,\xi): \xi = \mathbf{0} \}
\end{equation*}
is $e = \mathbf{0}$, $\xi = \mathbf{0}$ and $\Gamma(q)\tilde{\chi} = \mathbf{0}$.



Since $\dot{U}|_{\eqref{cls-I}} = -\xi^{T}K_{D}\xi \leq 0$, we can verify that
\begin{align*}
\frac{1}{2}K_{P}\|e(t)\|^{2} &\leq U(q(t),\tilde{\chi}(t),e(t),\xi(t)) \\
&\leq U(q(0),\tilde{\chi}(0),e(0),\xi(0)),~~ \forall t\geq 0.
\end{align*}
Hence, if the initial conditions of system \eqref{cls-I} satisfies $U(q(0),\tilde{\chi}(0),e(0),\xi(0)) < \frac{1}{2}K_{P} r^{2}$, then $\|e(t)\| < r$, $\forall t \geq 0$, i.e., the solution $e(t)$ of \eqref{cls-I} remains in the set $\mathcal{S}_{W_{r}}$, where $J(q)$ is always full rank as provided in Assumption~\ref{assume-J}.

Finally, by La-Salle’s invariance principle, we can conclude the closed-loop system \eqref{cls-I} locally converges to the set $e = \mathbf{0}$, $\xi = \mathbf{0}$ and $\Gamma(q)\tilde{\chi} = \mathbf{0}$ .
The proof is complete.

\eproof
\vspace{0.2cm}

One can relate the local information properties of distributed control law presented in Theorem \ref{thm-I} with that of distance-based formation control for single- and double-integrators in existing literature. More precisely, it can be checked that the implementation of distributed controller in Theorem \ref{thm-I} will be based only on local information defined on local coordinate frame in each agent. This local coordinate frame aspect is illustrated in Fig. \ref{fig-robots}.

\begin{remark}
In this paper, we have considered a team of perfect copies of manipulators, i.e., all the system parameters are exactly known, and there is no measurement noise. However, imperfections in sensing/parameters can introduce significant issues. For instance, system parameters can be slightly different from their nominal values, joint position measurement may have constant drift due to bias in the absolute encoder sensors, and there can be considerable amount of noise in joint velocity measurement. These undesirable factors may destabilize the  formation and introduce undesirable group motion. In this case, one can consider the use of adaptive control to compensate these uncertainties, as pursued in \cite{Marina2015controlling}. 
\end{remark}

\begin{table}[ht]
  \caption{\small The numerical values of the manipulator parameters used in the simulation. }\label{table-2DOF-para}
  \centering{\footnotesize
  \begin{tabular}{cccc}
    \toprule
    \multirow{2}{*}{Symbol} & \multirow{2}{*}{Meaning} & \multicolumn{2}{c}{$i$th link value} \\
    \cmidrule(r){3-4}
    & & $1$ & $2$ \\
    \midrule
    $m_{i}$ (Kg) & mass of $i$th link & $1.2$ & $1.0$ \\
    \midrule
    $I_{ci}$ (Kg$\cdot$m$^{2}$) & moments of inertia of $i$th link & 0.2250 & 0.1875 \\
    \midrule
    $l_{i}$ (m) & length of $i$th link & 1.5 & 1.5 \\
    \midrule
    $l_{ci}$ (m) & \tabincell{c}{distance from the center of the \\ mass of the $i$th link to the $i$th joint} & 0.75 & 0.75 \\
    \bottomrule
  \end{tabular}}
\end{table}


\begin{figure}
  \centering
  \includegraphics[width=0.48\textwidth]{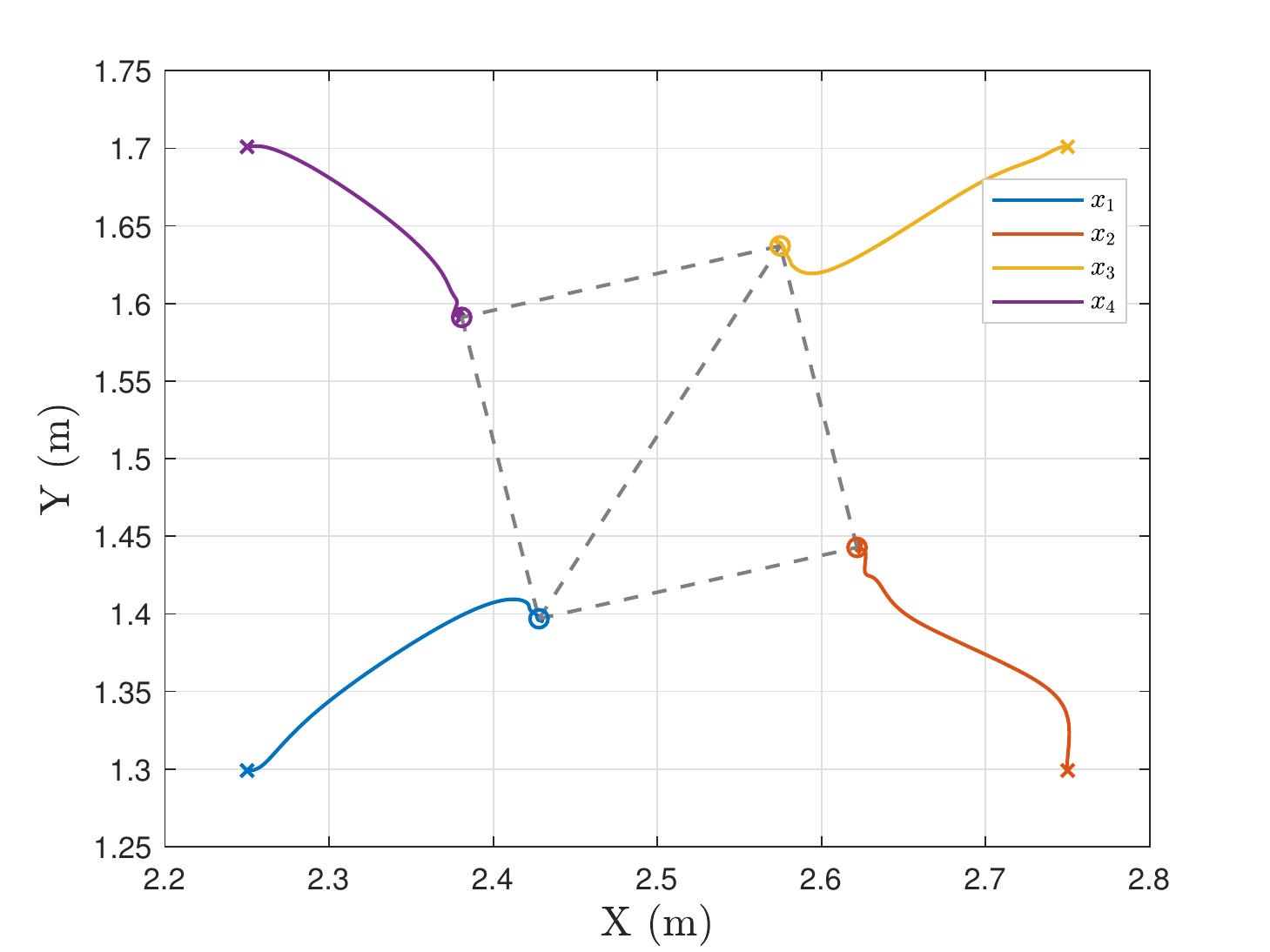}
  \caption{Trajectories of the manipulators' end-effector from the initial positions ($\times$) to the final positions ($\circ$).}\label{fig-2dof-XY}
\end{figure}

\begin{figure}
  \centering
  \includegraphics[width=0.48\textwidth]{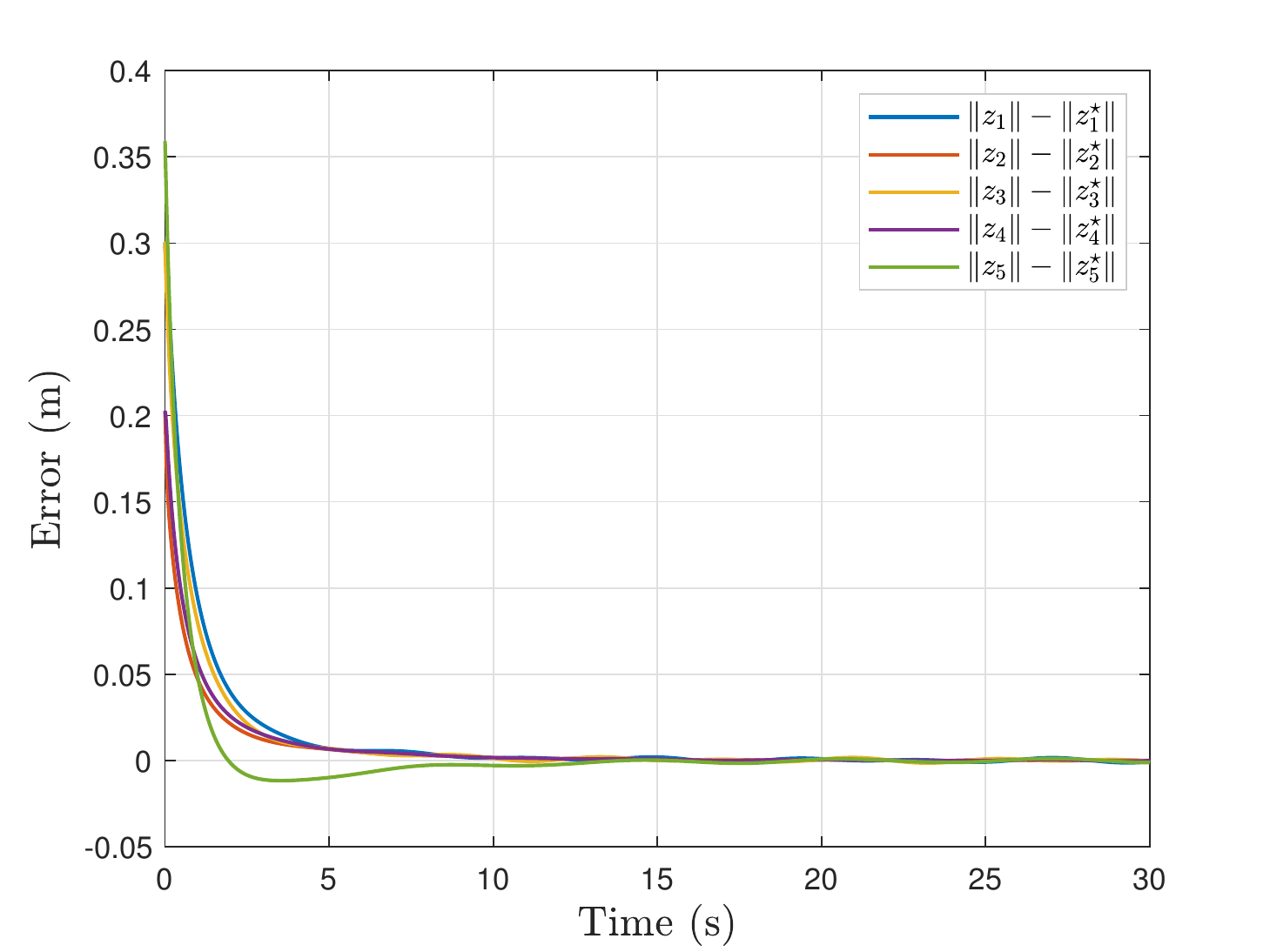}
  \caption{Performance of the inter-agents' distance errors.}\label{fig-2dof-e}
\end{figure}

\begin{figure}
  \centering
  \includegraphics[width=0.48\textwidth]{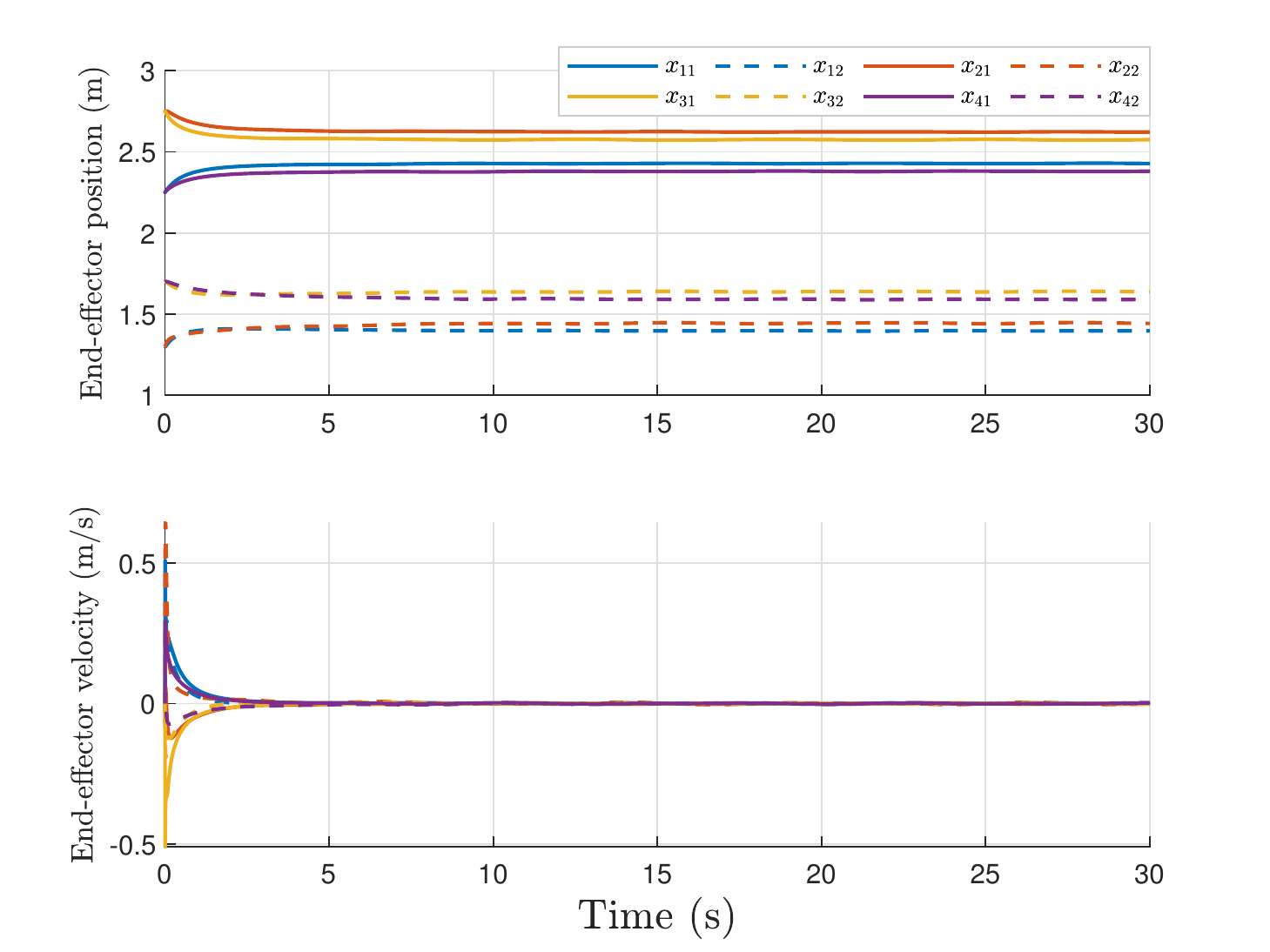}
  \caption{Performance of all the end-effector trajectories and velocities.}\label{fig-2dof-x}
\end{figure}

\begin{figure}
  \centering
  \includegraphics[width=0.48\textwidth]{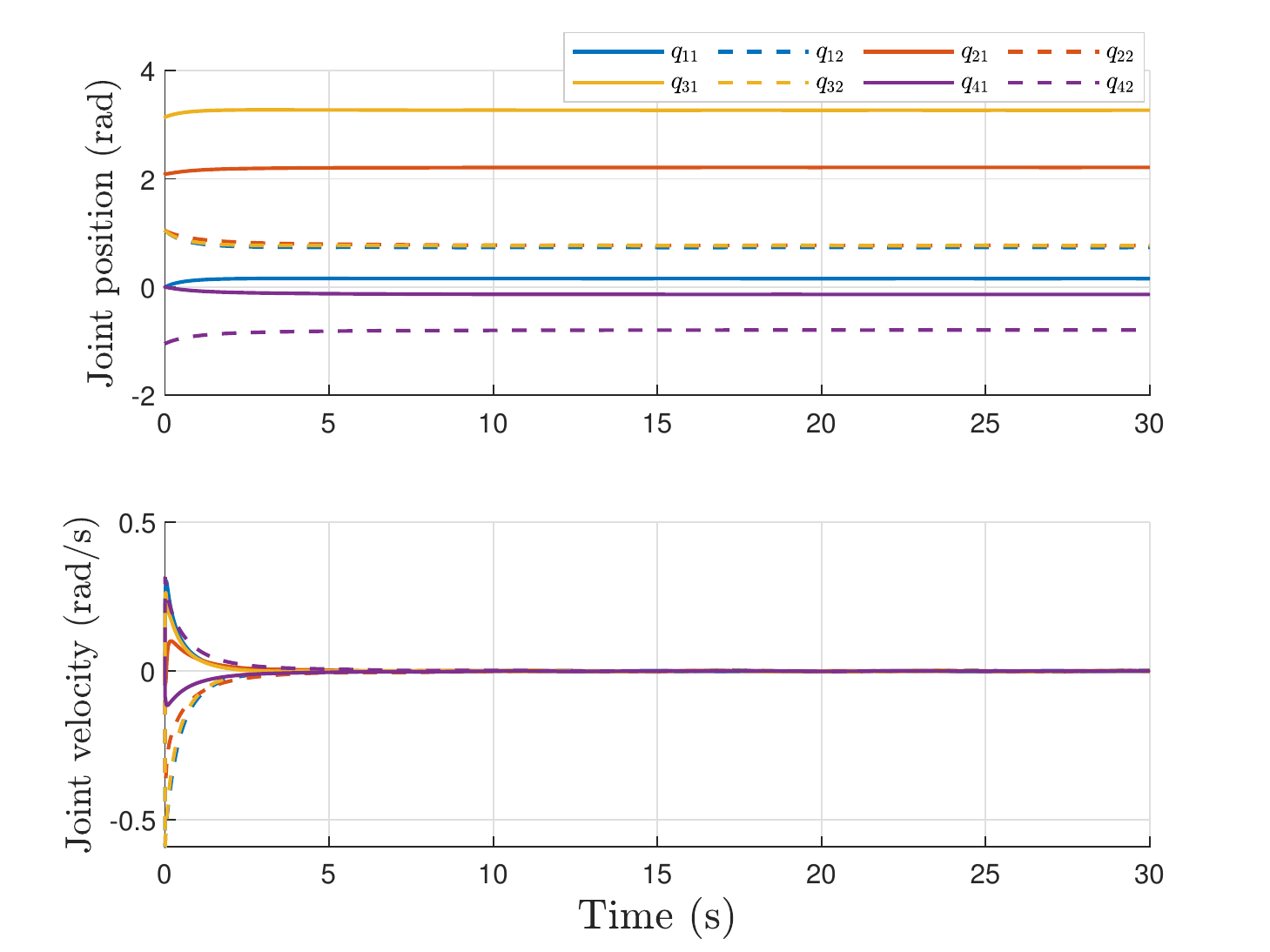}
  \caption{Performance of all the joint trajectories and velocities.}\label{fig-2dof-q}
\end{figure}

\section{SIMULATION}
We validate the design of distributed formation controller in Theorem \ref{thm-I} in this section by means of simulation.
For the simulation setup, we consider a network of $N=4$ two-link planar manipulator in the horizontal X-Y plane.
For the dynamic model of two-link robot manipulator as in \eqref{sys-dyna}, we refer to \cite[Example 6.2]{Slotine-book} and the corresponding numerical values of the parameters are 
given in Table~\ref{table-2DOF-para} for each link. 
The disturbances 
\eqref{defn-d} are set as $d_{M,i} = \bbm{\sin(t) & \sin(t)}^{T}$ and $d_{E,i} = \bbm{\frac{1}{2}\sin(\frac{\pi}{2}t) & \frac{1}{2}\sin(\frac{\pi}{2}t)}^{T}$ for $i=1,2,3,4$.
The kinematic model of each two-link robot manipulator is given by
\begin{align*}
x_{i} = \bbm{l_{1}\cos(q_{i1}) + l_{1}\cos(q_{i1} + q_{i2}) \\ l_{1}\sin(q_{i1}) + l_{1}\sin(q_{i1} + q_{i2})} + x_{i0}
\end{align*}
and correspondingly, its Jacobian matrix is given by
\begin{align*}
&J_{i}(q_{i}) \\
&= \bbm{-l_{1}\sin(q_{i1}) - l_{1}\sin(q_{i1} + q_{i2}) & -l_{1}\sin(q_{i1} + q_{i2}) \\ l_{1}\cos(q_{i1}) + l_{1}\cos(q_{i1} + q_{i2}) & l_{1}\cos(q_{i1} + q_{i2})}
\end{align*}
for $i=1,2,3,4$, where $q_{i} = \bbm{q_{i1} & q_{i2}}^{T}$.

We consider the formation shape of a square with side length of $0.4$ m and the associated  formation graph is represented by its incidence matrix given by  
\begin{equation*}
B = \begin{bmatrix} 1 & 0 & 0 & -1 & 1 \\ -1 & 1 & 0 & 0 & 0 \\ 0 & -1 & 1 & 0 & -1 \\ 0 & 0 & -1 & 1& 0 \end{bmatrix},
\end{equation*}
and illustrated in  Fig.~\ref{fig-robots} (right). 
For simulation setup of the manipulators, the bases of the 4 manipulators are located at $(0,0)$, $(5,0)$, $(5,3)$ and $(0,3)$, respectively and
the initial joint positions are set to $q_{1}=\bbm{0 &\pi/3}^{T}$, $q_{2}=\bbm{2\pi/3 & \pi/3}^{T}$, $q_{3}=\bbm{\pi & \pi/3}^{T}$, $q_{4}=\bbm{0 & -\pi/3}^{T}$. All the initial states of joint velocities and internal models are set to zero.

\addtolength{\textheight}{-3cm}

Using the distributed formation control as presented in  Theorem~\ref{thm-I}, we set the controller parameters as follows: 
$K_{P}=800$, $K_{D}=600$, $A_{M,i} = \text{block diag}\{[0~ 1; -1~ 0], [0~ 1; -1~ 0]$, $A_{E,i} = \text{block diag}\{[0~ \pi/2; -\pi/2~ 0], [0~ \pi/2; -\pi/2~ 0]\}$, $\Gamma_{M,i} = \text{block diag}\{[1~ 0], [1~ 0]\}$, and  $\Gamma_{E,i} = \text{block diag}\{[1~ 0], [1~ 0]\}$ for $i=1,2,3,4$.

Based on this simulation setup, we run the simulation for $30$s until the formation converges and the simulation results are shown in Figures 2 to 5. The trajectories and formation pattern of the manipulators' end-effector as presented 
in Fig.~\ref{fig-2dof-XY}. Fig.~\ref{fig-2dof-e} shows that the inner distance errors converge to zero as expected.  Fig.~\ref{fig-2dof-x}, which is the plot of end-effector positions and velocities, demonstrates clearly that the  formation of end-effectors towards the desired shape is achieved.  From Fig.~\ref{fig-2dof-q}, where the joint positions and velocities are plotted, we can conclude that the end-effectors remain stationary once they reach the intended shape, e.g., they do not exhibit undesirable group motion.

\section{CONCLUSIONS}
We have presented and analyzed  gradient descent-based distributed formation controllers for end-effectors which contains an internal model-based compensator to reject external disturbances.
The developed controller guarantees local asymptotic convergence to desired formation shape in spite of both input disturbance torques and disturbance forces at each end-effector. The efficacy of the proposed methods is shown in simulation.






\bibliographystyle{IEEEtran}
\bibliography{robotBIB}

\begin{thebibliography}{10}
\providecommand{\url}[1]{#1}
\csname url@samestyle\endcsname
\providecommand{\newblock}{\relax}
\providecommand{\bibinfo}[2]{#2}
\providecommand{\BIBentrySTDinterwordspacing}{\spaceskip=0pt\relax}
\providecommand{\BIBentryALTinterwordstretchfactor}{4}
\providecommand{\BIBentryALTinterwordspacing}{\spaceskip=\fontdimen2\font plus
\BIBentryALTinterwordstretchfactor\fontdimen3\font minus
  \fontdimen4\font\relax}
\providecommand{\BIBforeignlanguage}[2]{{%
\expandafter\ifx\csname l@#1\endcsname\relax
\typeout{** WARNING: IEEEtran.bst: No hyphenation pattern has been}%
\typeout{** loaded for the language `#1'. Using the pattern for}%
\typeout{** the default language instead.}%
\else
\language=\csname l@#1\endcsname
\fi
#2}}
\providecommand{\BIBdecl}{\relax}
\BIBdecl

\bibitem{Oh2015survey}
K.-K. Oh, M.-C. Park, and H.-S. Ahn, ``A survey of multi-agent formation
  control,'' \emph{Automatica}, vol.~53, pp. 424--440, 2015.

\bibitem{chan2021}
N.~Chan, B.~Jayawardhana, and H.~G. de~Marina, ``Angle-constrained formation
  control for circular mobile robots,'' \emph{IEEE Control Systems Letters},
  vol.~5, no.~1, pp. 109--114, 2021.

\bibitem{Oh2014distance}
K.-K. Oh and H.-S. Ahn, ``Distance-based undirected formations of
  single-integrator and double-integrator modeled agents in $n$-dimensional
  space,'' \emph{International Journal of Robust and Nonlinear Control},
  vol.~24, no.~12, pp. 1809--1820, 2014.

\bibitem{Marina2015controlling}
H.~G. {de Marina}, M.~Cao, and B.~Jayawardhana, ``Controlling rigid formations
  of mobile agents under inconsistent measurements,'' \emph{IEEE Transactions
  on Robotics}, vol.~31, no.~1, pp. 31--39, 2015.

\bibitem{Marina2018taming}
H.~G. {de Marina}, B.~Jayawardhana, and M.~Cao, ``Taming mismatches in
  inter-agent distances for the formation-motion control of second-order
  agents,'' \emph{IEEE Transactions on Automatic Control}, vol.~63, no.~2, pp.
  449--462, 2018.

\bibitem{zhao2015bearing}
S.~Zhao and D.~Zelazo, ``Bearing rigidity and almost global bearing-only
  formation stabilization,'' \emph{IEEE Transactions on Automatic Control},
  vol.~61, no.~5, pp. 1255--1268, 2015.

\bibitem{Jafarian2016disturbance}
M.~Jafarian, E.~Vos, C.~{De Persis}, J.~M.~A. Scherpen, and A.~{van der
  Schaft}, ``Disturbance rejection in formation keeping control of nonholonomic
  wheeled robots,'' \emph{International Journal of Robust and Nonlinear
  Control}, vol.~26, no.~15, pp. 3344--3362, 2016.

\bibitem{vos2016formation}
E.~Vos, A.~J. {van der Schaft}, and J.~M.~A. Scherpen, ``Formation control and
  velocity tracking for a group of nonholonomic wheeled robots,'' \emph{IEEE
  Transactions on Automatic Control}, vol.~61, no.~9, pp. 2702--2707, 2016.

\bibitem{Scharf2003survey}
D.~P. Scharf, F.~Y. Hadaegh, and S.~R. Ploen, ``A survey of spacecraft
  formation flying guidance and control (part {I}): {Guidance},'' in
  \emph{Proceedings of the 2004 American control conference}, 2003, pp.
  1733--1739.

\bibitem{Scharf2004survey}
------, ``A survey of spacecraft formation flying guidance and control (part
  {II}): {Control},'' in \emph{Proceedings of the 2004 American control
  conference}, 2004, pp. 2976--2985.

\bibitem{Xu2017formation}
D.~Xu, X.~Wang, Y.~Su, and D.~Wang, ``Formation control in dynamic positioning
  of multiple offshore vessels via cooperative robust output regulation,'' in
  \emph{2017 IEEE 56th Annual Conference on Decision and Control (CDC)}.\hskip
  1em plus 0.5em minus 0.4em\relax IEEE, 2017, pp. 4070--4075.

\bibitem{Chen1997adaptive}
B.~S. Chen, Y.~C. Chang, and T.~C. Lee, ``Adaptive control in robotic systems
  with ${H}_{\infty}$ tracking performance,'' \emph{Automatica}, vol.~33,
  no.~2, pp. 227--234, 1997.

\bibitem{Bayu2008}
B.~Jayawardhana and G.~Weiss, ``Tracking and disturbance rejection for fully
  actuated mechanical systems,'' \emph{Automatica}, vol.~44, no.~11, pp.
  2863--2868, 2008.

\bibitem{Lu2019}
M.~Lu, L.~Liu, and G.~Feng, ``Adaptive tracking control of uncertain
  {Euler–Lagrange} systems subject to external disturbances,''
  \emph{Automatica}, vol. 104, pp. 207--219, 2019.

\bibitem{Wu2019ICCA}
H.~Wu and D.~Xu, ``Inverse optimality and adaptive asymptotic tracking control
  of uncertain {Euler-Lagrange} systems,'' in \emph{2019 IEEE 15th
  International Conference on Control and Automation (ICCA)}, 2019, pp.
  242--247.

\bibitem{Verginis2019cooperative}
C.~K. Verginis, D.~Zelazo, and D.~V. Dimarogonas, ``Cooperative manipulation
  via internal force regulation: {A} rigidity theory perspective,'' \emph{arXiv
  preprint arXiv:1911.01297}, 2019.

\bibitem{Ren2020fully}
Y.~Ren, S.~Sosnowski, and S.~Hirche, ``Fully distributed cooperation for
  networked uncertain mobile manipulators,'' \emph{IEEE Transactions on
  Robotics}, vol.~36, no.~4, pp. 984--1003, 2020.

\bibitem{Dohmann2020distributed}
P.~B. {gen. Dohmann} and S.~Hirche, ``Distributed control for cooperative
  manipulation with event-triggered communication,'' \emph{IEEE Transactions on
  Robotics}, vol.~36, no.~4, pp. 1038--1052, 2020.

\bibitem{Slotine-book}
J.~J.~E. Slotine and W.~Li, \emph{Applied Nonlinear Control}.\hskip 1em plus
  0.5em minus 0.4em\relax Englewood Cliffs, NJ: Prentice hall, 1991.

\bibitem{Murray1994}
R.~M. Murray, Z.~Li, and S.~S. Sastry, \emph{A Mathematical Introduction to
  Robotic Manipulation}.\hskip 1em plus 0.5em minus 0.4em\relax CRC press,
  1994.

\bibitem{Spong-book}
M.~W. Spong, S.~Hutchinson, and M.~Vidyasagar, \emph{Robot Modeling and
  Control}.\hskip 1em plus 0.5em minus 0.4em\relax New York: Wiley, 2006.

\bibitem{Ortega-book}
R.~Ortega, A.~Lor\'{\i}a, P.~J. Nicklasson, and H.~Sira-Ram\'{\i}rez,
  \emph{Passivity-based Control of Euler-Lagrange Systems: Mechanical,
  Electrical and Electromechanical Applications}.\hskip 1em plus 0.5em minus
  0.4em\relax Springer Science \& Business Media, 1998.

\bibitem{Kelly-book}
R.~Kelly, V.~S. Davila, and A.~Lor\'{\i}a, \emph{Control of Robot Manipulators
  in Joint Space}.\hskip 1em plus 0.5em minus 0.4em\relax Springer Science \&
  Business Media, 2005.

\bibitem{de2020maneuvering}
H.~G. de~Marina, ``Maneuvering and robustness issues in undirected
  displacement-consensus-based formation control,'' \emph{IEEE Transactions on
  Automatic Control}, 2020.

\bibitem{de2016distributed}
H.~G. {de Marina}, B.~{Jayawardhana}, and M.~Cao, ``Distributed rotational and
  translational maneuvering of rigid formations and their applications,''
  \emph{IEEE Transactions on Robotics}, vol.~32, no.~3, pp. 684--697, 2016.

\bibitem{AnYuFiHe08}
B.~D.~O. Anderson, C.~Yu, B.~Fidan, and J.~Hendrickx, ``Rigid graph control
  architectures for autonomous formations,'' \emph{IEEE Control Systems
  Magazine}, vol.~28, pp. 48--63, 2008.

\bibitem{van2014pH}
A.~{van der Schaft} and D.~Jeltsema, ``{Port-Hamiltonian} systems theory: {An}
  introductory overview,'' \emph{Foundations and Trends in Systems and
  Control}, vol.~1, no. 2-3, pp. 173--378, 2014.

\bibitem{Cheah2003approximate}
C.~C. Cheah, M.~Hirano, S.~Kawamura, and S.~Arimoto, ``Approximate jacobian
  control for robots with uncertain kinematics and dynamics,'' \emph{IEEE
  Transactions on Robotics and Automation}, vol.~19, no.~4, pp. 692--702, 2003.

\bibitem{Dixon2007adaptive}
W.~E. Dixon, ``Adaptive regulation of amplitude limited robot manipulators with
  uncertain kinematics and dynamics,'' \emph{IEEE Transactions on Automatic
  Control}, vol.~52, no.~3, pp. 488--493, 2007.

\bibitem{Wang2020dynamic}
H.~Wang, W.~Ren, C.~C. Cheah, Y.~Xie, and S.~Lyu, ``Dynamic modularity approach
  to adaptive control of robotic systems with closed architecture,'' \emph{IEEE
  Transactions on Automatic Control}, vol.~65, no.~6, pp. 2760--2767, 2020.

\bibitem{Khalil2002}
H.~K. Khalil, \emph{Nonlinear Systems}.\hskip 1em plus 0.5em minus 0.4em\relax
  New Jersey: Prentice Hall, 2002.

\end{thebibliography}

\end{document}